\documentclass[final,times,12pt,numbers]{elsarticle}

\usepackage{amssymb}
\usepackage{bm}

\usepackage{hyperref}
\hypersetup{
    colorlinks=true,
    linkcolor=blue,
    filecolor=magenta,
    urlcolor=cyan,
    pdfpagemode=FullScreen
}
\usepackage{makecell}

\usepackage{subcaption}

\journal{Image and Vision Computing}

\begin{document}

\begin{frontmatter}



\title{Sign language recognition based on deep learning and low-cost handcrafted descriptors}


\author{Alvaro Leandro Cavalcante Carneiro}
\ead{alvaro.leandro@unesp.br}

\author{Denis Henrique Pinheiro Salvadeo}
\ead{denis.salvadeo@unesp.br}

\author{Lucas de Brito Silva}
\ead{lucas.brito-silva@unesp.br}

\affiliation{organization={São Paulo State University, Institute of Geosciences and Exact Sciences},
            addressline={Av. 24 A}, 
            city={Rio Claro},
            postcode={13506900}, 
            state={São Paulo},
            country={Brazil}}

\begin{abstract}

In recent years, deep learning techniques have been used to develop sign language recognition systems,
potentially serving as a communication tool for millions of hearing-impaired individuals worldwide. However, there are inherent challenges in creating such systems. Firstly, it is important to consider as many linguistic parameters as possible in gesture execution to avoid ambiguity between words. Moreover, to facilitate the real-world adoption of the created solution, it is essential to ensure that the chosen technology is realistic, avoiding expensive, intrusive, or low-mobility sensors, as well as very complex deep learning architectures that impose high computational requirements. Based on this, our work aims to propose an efficient sign language recognition system that utilizes low-cost sensors and techniques. To this end, an object detection model was trained specifically for detecting the interpreter's face and hands, ensuring focus on the most relevant regions of the image and generating inputs with higher semantic value for the classifier. Additionally, we introduced a novel approach to obtain features representing hand location and movement by leveraging spatial information derived from centroid positions of bounding boxes, thereby enhancing sign discrimination. The results demonstrate the efficiency of our handcrafted features, increasing accuracy by 7.96\% on the AUTSL dataset, while adding fewer than 700 thousand parameters and incurring less than 10 milliseconds of additional inference time. These findings highlight the potential of our technique to strike a favorable balance between computational cost and accuracy, making it a promising approach for practical sign language recognition applications.
\end{abstract}


\end{frontmatter}


\section{Introduction}
\label{introduction}

According to the World Health Organization \cite{who}, approximately 430 million people worldwide experience severe hearing loss. This condition, along with deafness, significantly affects individuals' quality of life and poses challenges to their social integration. Despite sign language being a widely used form of communication within the deaf community, a huge barrier exists as the majority of people are not familiar with this language.

A possible solution for this problem is the development of sign language recognition systems, capable of translating visual signs into subtitles in the users' native language, becoming a valuable communication assistance tool for the hearing impaired and promoting accessibility. The advances in deep learning and computer vision are enabling the creation of highly accurate models, as shown by previous research in this field \cite{de2021isolated, de2020sign, gruber2021mutual, sincan2021using, gokcce2020score}. Nevertheless, we are yet to achieve a solution that can be effectively used by deaf people in their daily routines, primarily due to limitations in portability and performance of existing recognition systems. It usually happens because many of the authors rely on non-portable sensors to capture the sign, such as Leap Motion \cite{potter2013leap} or Kinect \cite{han2013enhanced}, while others employ very complex architectures \cite{belissen2019automatic} with high computational requirements, making them impractical to execute on portable devices.

Based on that, this work aims to better explore the linguistic parameters underlying sign formation and present a novel sign language recognition system that prioritizes efficiency and low computational cost. 

\begin{figure}[!ht]
    \centering
    \caption{Location of each of the linguistic parameters in sign language.}
    \includegraphics[width=.99\textwidth]{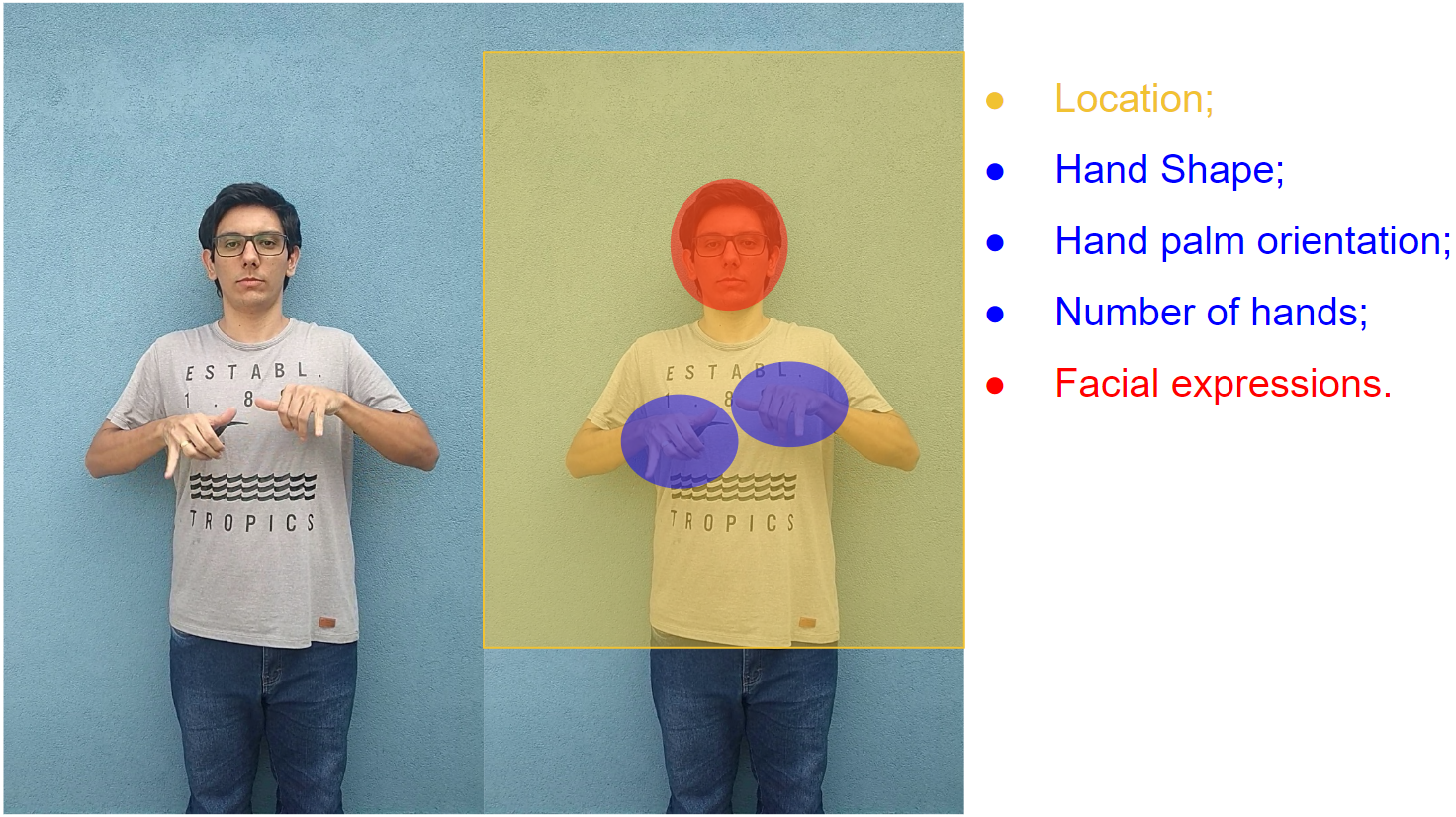}
    \label{fig:language_params}

    Source: Author.
\end{figure}

There are six fundamental linguistic parameters \cite{xavier2014barbosa} that contribute to the semantics of gestures in sign language, including hand shape, location, movement, hand palm orientation, number of hands, and facial expressions.  Figure \ref{fig:language_params} illustrates all of the parameters except for movement.

As highlighted by the figure, most part of the relevant information occupies a small portion of the image, implying that only some segments need to be considered for classification. Therefore, we employed an optimized and lightweight object detector in the preprocessing stage to segment the interpreters' hands and face, reducing the amount of data that requires processing during model training and inference. 

However, relying solely on the hand and face segments would lead to a loss of spatial reference, which is essential to represent the execution location of the sign in relation to the body, introducing ambiguity to some words and impacting the system performance. For that reason, our research also proposes a novel set of handcrafted features that are based on referential attributes extracted from the bounding box centroids generated by the object detector, in order to represent the movement and location parameters. By using the suggested features, we intend to replace the background and the interpreter's body with a simple vector representation, further reducing the computational cost. 

\begin{figure}[!ht]
    \centering
    \caption{Architecture overview of the proposed system.}
    \includegraphics[width=.99\textwidth]{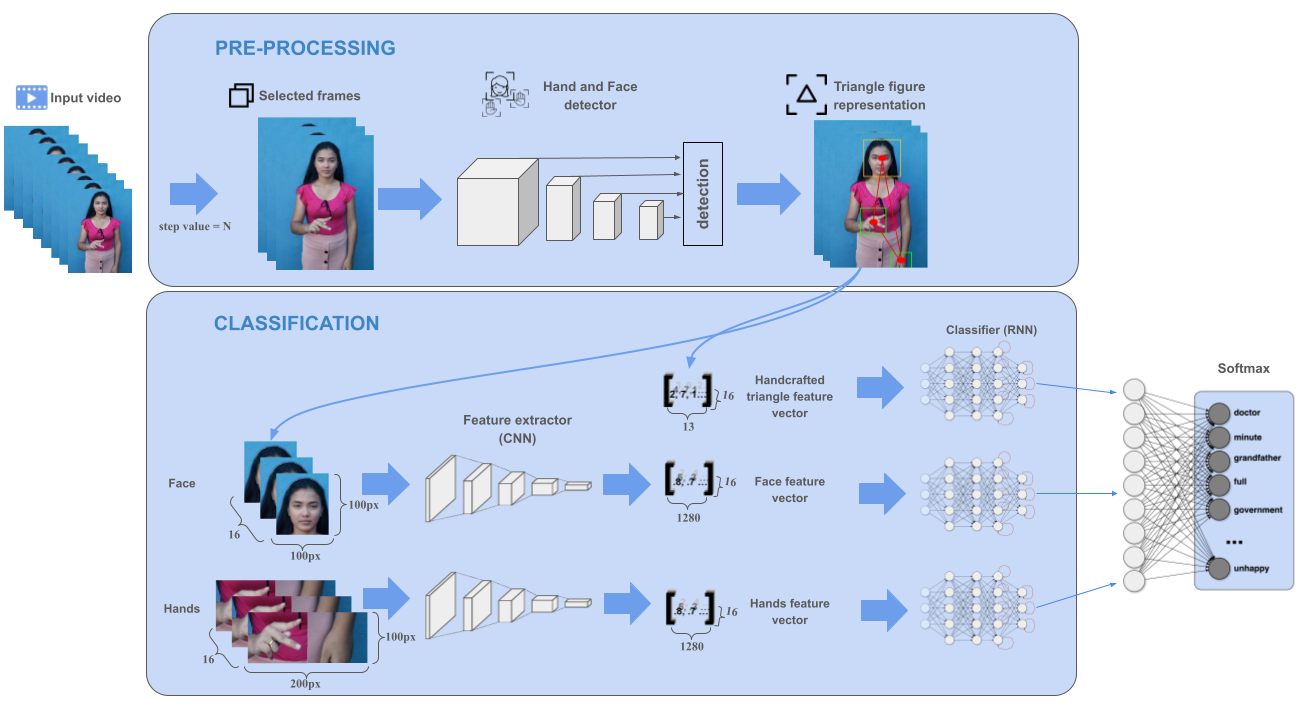}
    \label{fig:full_architecture}

    Source: Author.
\end{figure}

Besides that, we applied multiple techniques in the preprocessing of the videos to carefully subsample the frames, ensuring that each selected frame has a significant semantic value for sign classification. Finally, we employed a Convolutional Neural Network (CNN) to extract features from the images and different Recurrent Neural Network (RNN) models for each channel to generate the predictions. The architecture overview is presented in Figure \ref{fig:full_architecture}.

The main contributions of this paper are as follows:

\begin{itemize}
    \item A novel set of handcrafted features that are capable of representing hand location and movement, improving the model accuracy with low computational cost.
    \item A preprocessing stage focused on frame subsampling, extracting the maximum semantic value from a limited number of frames in the video.
\end{itemize}

\section{Related Works}\label{related_works}

Sign language recognition has been a subject of extensive research in recent years, leading to numerous approaches aimed at translating visual signs into text. Even within the subset of works that utilize RGB sensors, which are more suitable for real-world scenarios, there exists a wide variety of models and preprocessing strategies. Notably, object detection and body pose inference have emerged as prominent techniques, enabling the segmentation of important parts of the interpreter's body, in addition to allowing the creation of new relevant features for classification.

However, an important aspect that is usually ignored by previous research is the computational requirements of the proposed solution, which can directly affect the adoption of the system. In this regard, while body pose estimation is commonly utilized in the literature \cite{de2021isolated, de2020sign, gruber2021mutual, gokcce2020score, belissen2019automatic, kratimenos2021independent, maruyama2021word, pu2016sign}, it comes with a notable increase in computational cost. On the other hand, object detection offers the advantage of real-time performance \cite{wang2022yolov7}, even on edge devices \cite{martinez2022smartphone}.

Furthermore, our approach goes beyond merely segmenting specific parts of the image, as demonstrated in prior works \cite{liao2019dynamic, zhang2021sign, agrawal20202}. Our aim is to leverage the reference information provided by the bounding boxes centroids to introduce novel features. In this sense, there are only a few works that used a similar approach to the one presented. For instance, Talukdar e Bhuyan \citep{talukdar2022vision}, created a triangular reference between the face and hands, extracting new features such as the coordinates of the centroids, a movement vector based on centroids positions, a histogram of oriented gradients and the orientation of ellipses drawn around the hand centroids. Nevertheless, the purpose of their work was solely to distinguish between gestures that represent signs and transitory gestures, without applying the features for sign classification.

Xiao et al. \cite{xiao2020multi}, on the other hand, adopted an object detector to localize the hands and face and establish a triangular reference, similar to our approach. The primary distinction lies in their proposal of a triangular figure (image of the same input size with the triangle drawn on a constant background, i.e., a triangle edge mask) instead of a handcrafted feature vector. This figure was then subjected to a CNN for feature extraction and subsequently to an LSTM-type \cite{hochreiter1997long} RNN for classification. Although the authors' approach was effective, the use of a CNN significantly increases the computational cost of the system, while generating non-explainable features due to its black-box nature. Even so, we adopted this triangle-figure approach as a benchmark and baseline for our novel handcrafted features, enabling a direct comparison between the two techniques, as detailed in Section \ref{materials}.

\section{Materials and methods}
\label{materials}
The development of the proposed system encompassed multiple stages, ranging from dataset selection to architectural particularities of the classification model. The subsequent sections present comprehensive details regarding each of these steps.

\subsection{Dataset} \label{dataset}
Although our approach is generalist, we initially focused on word-level sign language recognition, where each video in the dataset contains a single sign. Considering this context, we chose to use the AUTSL (Ankara University Turkish Sign Language) \cite{sincan2020autsl} dataset due to its remarkable complexity, arising from the extensive range of interpreters, signs, and backgrounds it encompasses.

The selected dataset comprises 226 signs performed by 43 interpreters across 20 distinct scenarios. The recordings were captured at a resolution of 512x512 pixels, with a frame rate of 30 frames per second. In total, the dataset includes 38,336 video samples, with an average duration of approximately 2.05 seconds and 61 frames per video. 

To guarantee a proper dataset split, we followed the guidelines established by the ChaLearn competition \cite{sincan2021chalearn}. Accordingly, approximately 77.52\% of the data was allocated for training, 12.17\% for validation, and 10.31\% for testing. Furthermore, training, testing, and validation sets consist of samples from 31, 6, and 6 interpreters, respectively, ensuring that each interpreter appears in a single set.

\subsection{Object detection} \label{object detection}
Object detection plays a crucial role as it provides the necessary bounding boxes for feature engineering. To maximize the accuracy of the detector, we created our own hand and face detection dataset\footnote{\url{https://github.com/AlvaroCavalcante/hand-face-detector}} focused on sign language \cite{carneirolarge}, enabling to train a model that is specifically designed for this problem domain. To build the dataset, we carefully sub-sampled and annotated 477,480 frames from the AUTSL videos, ensuring that each image contains two visible hands and one face. In addition, we divided the dataset into separate subsets for training, testing, and validation based on the interpreter, as outlined in Section \ref{dataset}.

Furthermore, we employed the CenterNet \cite{zhou2019objects} architecture for object detection due to its favorable trade-off between computational cost and precision, as highlighted in our previous work \cite{carneirolarge}. 

\begin{figure}[!ht]
     \centering
     \caption{Data augmentation techniques for object detection.}
     \begin{subfigure}[t]{0.3\textwidth}
         \centering
         \includegraphics[width=\textwidth]{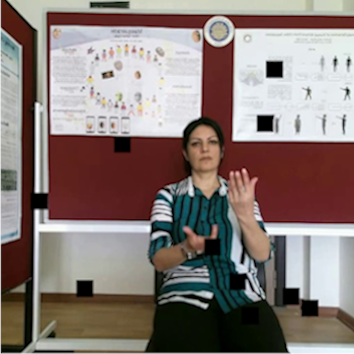}
         \caption{Small random black patches.}
         \label{fig:augmentation_detection_1}
     \end{subfigure}
     \hfill
     \begin{subfigure}[t]{0.3\textwidth}
         \centering
         \includegraphics[width=\textwidth]{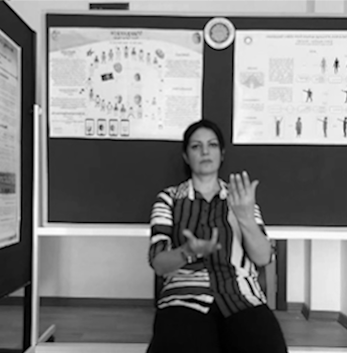}
         \caption{grayscale conversion.}
         \label{fig:augmentation_detection_2}
     \end{subfigure}
     \hfill
     \begin{subfigure}[t]{0.3\textwidth}
         \centering
         \includegraphics[width=\textwidth]{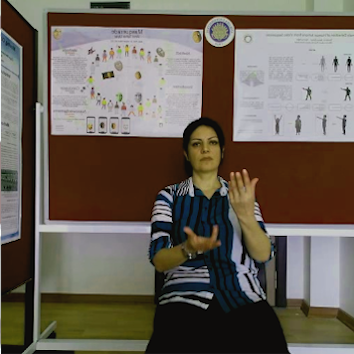}
         \caption{Color channel distortion.}
         \label{fig:augmentation_detection_3}
     \end{subfigure}

        Source: Author.
        \label{fig:augmentation_techniques}
\end{figure}

Finally, we diligently selected a set of data augmentation techniques to be applied during the training of the object detection model, including grayscale conversion, random distortion in RGB color channels, Hue changes, small random black patches, as well as adjustments in brightness, contrast, and saturation of the image. Moreover, geometric operations such as horizontal rotation, image scale variations, and random cropping were also applied. Figure \ref{fig:augmentation_techniques} provides a visual illustration of some of the aforementioned operations.

\subsection{Video pre-processing} \label{video preprocessing}
The objective of the pre-processing stage was to extract all the information channels that comprise the architecture, illustrated in Figure \ref{fig:full_architecture}. To this end, several methods and algorithms were implemented to ensure that the maximum amount of information is preserved for sign classification. During each iteration through the dataset, a batch of 180 videos was randomly selected and loaded into memory. Subsequently, only 16 frames from each video were used to represent the temporal information of the signs, as observed in previous work \cite{zhang2021sign}.

\begin{figure}[ht]
    \centering
    \caption{Use of the step value for frame subsampling.}
    \includegraphics[width=.99\textwidth]{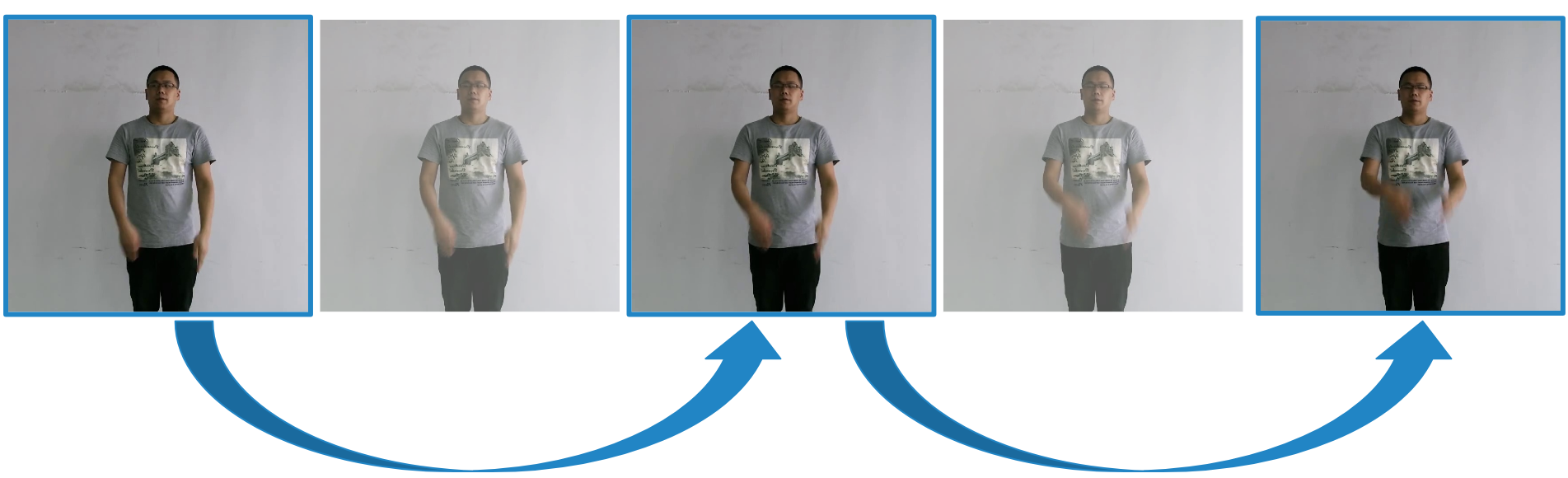}
    \label{fig:video_subsample}

    Source: Author.
\end{figure}

While increasing the size of the temporal sequence minimizes the loss of relevant information during frame subsampling, longer sequences can considerably increase the complexity of training recurrent networks \cite{geron2019hands} and result in higher computational cost for generating model inferences. Therefore, the subsampling process was carefully optimized to retain the maximum amount of information in just 16 frames. To achieve this, we first defined a step value that was used as a sampling criterion, obtained by dividing the total number of frames in the video by 16. For example, a video with 39 frames would result in a step of 2, rounded to the lowest value. This process ensures equally spaced sampling by time, as illustrated in Figure \ref{fig:video_subsample}.

\begin{figure}[!ht]
    \centering
    \caption{Frames from different time intervals (indicated by numbers in red) from one of the AUTSL videos.}
    \includegraphics[width=.99\textwidth]{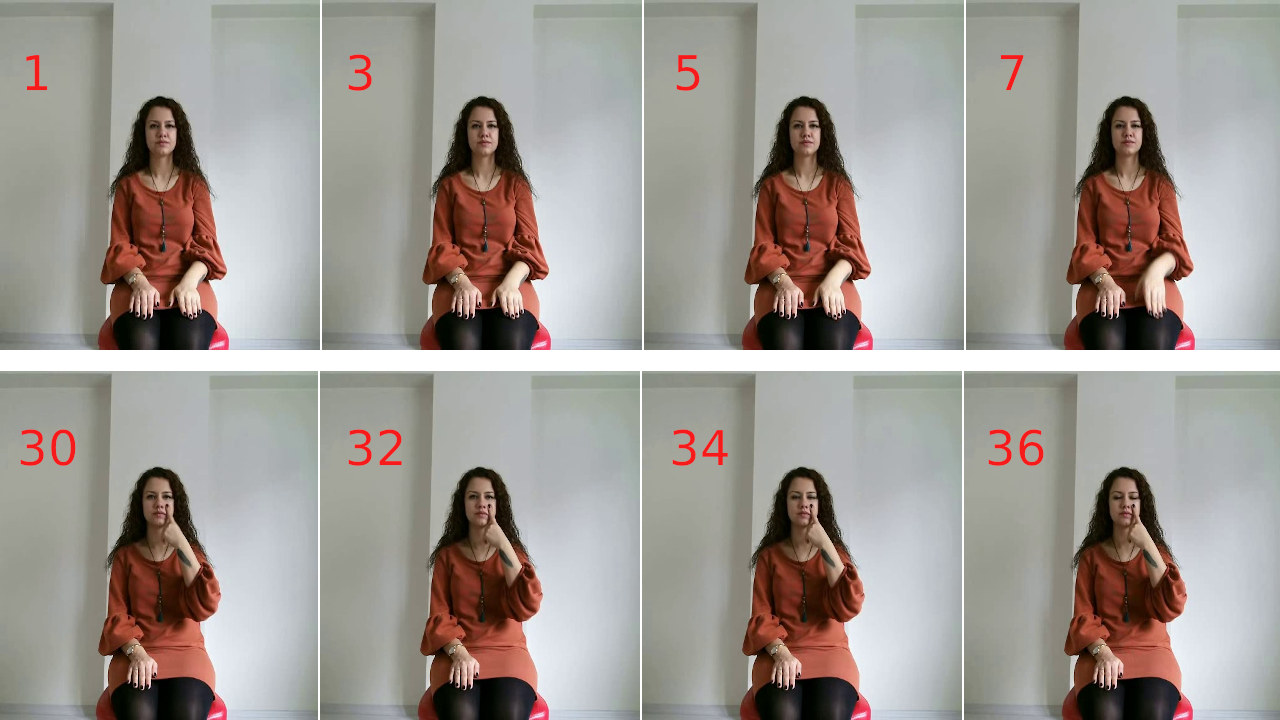}
    \label{fig:sign_problems}
 
    Source: Author.
\end{figure}

However, several problems were observed during the subsampling of the data. First, the initial and final frames of the videos usually depict the resting pose, which has no discriminative information. Additionally, signs executed slowly by the interpreter or those with longer execution sequences often result in a significant number of similar frames. This occurs because the interpreter tends to maintain static hand positions during transitions between signs or at the end of the sign's execution. Figure \ref{fig:sign_problems} illustrates the initial frames (on top), and the frames during the execution of the sign (on bottom) of one of the videos from the AUTSL dataset, highlighting the aforementioned problems.

To address these challenges, a criterion was established to filter and select the most informative video frames based on the hand movement history during the execution of the sign. Using this approach, a position value $\rho$ was calculated for each of the selected frames, as defined in the Equation \ref{eq:position}:

\begin{equation} \label{eq:position}
   \rho = d_{2} + d_{3}.
\end{equation}

Due to the interpreter's head being a fixed point during sign execution, the distance between the hands and the face ($d_{2}$ and $d_{3}$) can serve as a reference for spatial positioning, as detailed in Section \ref{handcrafted features}. The positions are stored in a vector $\mathbf{p}$, representing the history of hand positions throughout the video. From the second frame onward, a $\mu_{t}$ value is calculated to measure movement, as shown in Equation \ref{eq:movimento}:

\begin{equation} \label{eq:movimento}
   \mu_{t} = \big| p_{t} - p_{t-1} \big|,
\end{equation}

\noindent where $p_{t}$ represents the position of the hands at time $t$, and $p_{t-1}$ is the position in the previous frame of the video at time $t-1$. The movement information represents the degree of displacement of the hands between consecutive frames, and it is stored in a vector $\bm{\mu}$ that was used as a criterion for selecting the best frames.

Regarding the resting pose issue, from the first frame of the video, all subsequent frames are discarded until the hand movement exceeds a threshold value $\eta$. Additionally, after overcoming the resting pose, the last three time intervals of the motion vector are evaluated at each iteration. If all $\mu_{t}$ values are less than a threshold value $\hat{\eta}$, the frame is discarded. This approach addresses the problem caused by slow transitions by eliminating consecutive frames that exhibit very little movement and are, consequently, similar to each other. The values of $\eta$ and $\hat{\eta}$ are related to the Euclidean distance and were empirically defined as 10 and 5, respectively, improving the sampling of frames and avoiding redundant information.

\begin{figure}[ht]
    \centering
    \caption{Comparison of model-generated bounding boxes (in red) with manually increased size by 10\% (in green).}
    \includegraphics[width=.99\textwidth]{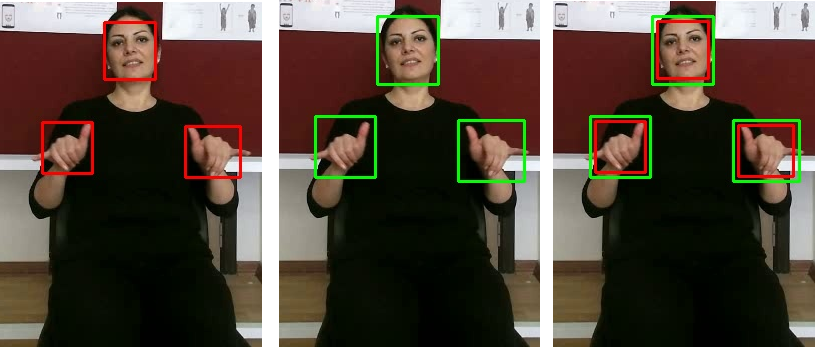}
    \label{fig:aumento_crop_bbox}

    Source: Author.
\end{figure}

Concerning object detection, a confidence threshold of 55\% was used to accept the inferences, and the face and the two hands with the highest confidence scores were selected, minimizing the chances of incorporating noisy segments into the dataset. Moreover, to prevent the interpreter's fingers from being cropped, the area of the bounding boxes generated by the model was increased by 10\%, as illustrated in Figure \ref{fig:aumento_crop_bbox}.

However, in cases where the model fails to detect the non-dominant hand or the face in a frame at instant $t$, the detections from the previous instant are used to fill in the missing information. Conversely, failures in detecting the dominant hand are not tolerated, as this hand holds a more significant semantic value, leading to the exclusion of the frame.

\begin{figure}[!ht]
     \centering
     \caption{Process used to differentiate the hands.}
     \begin{subfigure}[t]{0.3\textwidth}
         \centering
         \includegraphics[width=\textwidth]{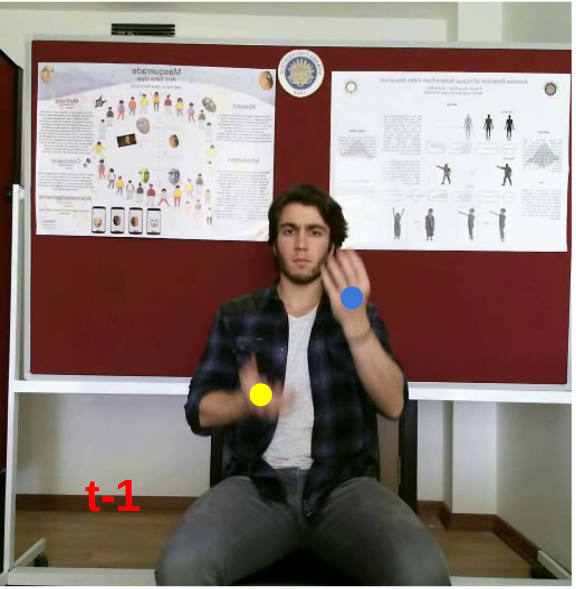}
         \caption{Centroids detected in the previous time interval, distinguished between ``hand 1'' (yellow) and ``hand 2'' (blue).}
         \label{fig:centroid_fig1}
     \end{subfigure}
     \hfill
     \begin{subfigure}[t]{0.3\textwidth}
         \centering
         \includegraphics[width=\textwidth]{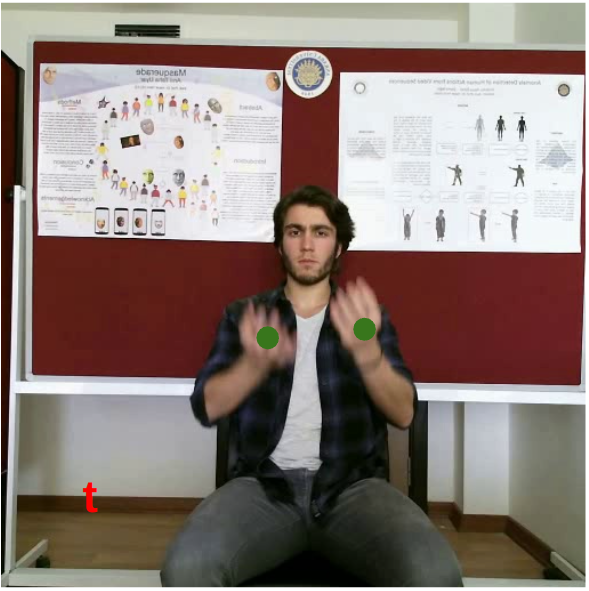}
         \caption{Centroids in the current time, without distinction.}
         \label{fig:centroid_fig2}
     \end{subfigure}
     \hfill
     \begin{subfigure}[t]{0.3\textwidth}
         \centering
         \includegraphics[width=\textwidth]{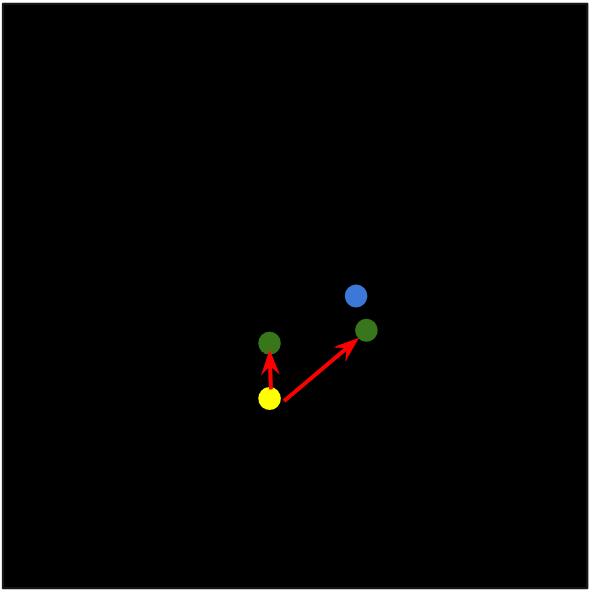}
         \caption{Calculation of the distance between centroids to identify hands based on proximity.}
         \label{fig:centroid_fig3}
     \end{subfigure}

        Source: Author.
        \label{fig:mao_dominante}
\end{figure}

To enable this differentiation, the hands were categorized as either ``hand 1'' or ``hand 2'' based on the smallest Euclidean distance between the centroids at time $t$ and $t-1$. Specifically, after generating object detector inferences at time $t$, the distance between the centroid of ``hand 1'' (obtained at $t-1$) and the two new hand centroids in the current time interval is computed. The new closest centroid is then classified as ``hand 1'', while the farthest one is labeled as ``hand 2'', as illustrated in Figure \ref{fig:mao_dominante}. In the first frame of the video, where previous detections are not available, hands are categorized arbitrarily.

Once that this is done, a motion vector is generated for each hand based on its position history, as discussed earlier. After the frame subsampling, we classify the hand with the highest motion value across all selected frames as dominant.

Finally, the aforementioned criteria may result in the exclusion of a substantial number of frames from certain videos, leading to the sampling of fewer than 16 frames. In such cases, the iteration through the video is restarted, reducing the step value by 1. However, even considering this strategy, it is possible that the minimum of 16 frames is not achieved, especially in shorter or noisy videos. In that case, the missing data is addressed through zero-padding. As a final validation measure, videos containing more than 5 out of the 16 frames filled with zeros are discarded due to the lack of information.

\subsection{Data augmentation} \label{data augmentation}
Data augmentation is of paramount importance in the context of image classification, as it promotes model invariance. Following this trend, we randomly applied various transformations to the image segments of the hands and face. These operations were dynamically implemented during the model training iterations, with each transformation having a 50\% probability of being applied. Moreover, the images were subject to the possibility of undergoing multiple concurrent transformations.

Among the geometric operations, we applied rotation, shear, zoom and horizontal and vertical shift. Besides that, 
to prevent any unintended alterations to the semantics of the signs, we carefully optimized the range of application for each transformation, as detailed in Table \ref{table:data_aug_ranges}. Additionally, zoom and shift were intentionally excluded from the hand segment to retain the integrity of the interpreter's finger information.

\begin{table}[h!]
\centering
\caption{Ranges of application of the data augmentation techniques.}
\label{table:data_aug_ranges}
\vspace*{.25 cm}

 \begin{tabular}{| c c|} 
 \hline
 \textbf{Operation} & \textbf{Range} \\ [0.5ex] 
 \hline\hline
 Rotation & -20°-- 20° \\ 
 \hline
 Shear & 5°-- 12° \\ 
 \hline
 Zoom & 0.8 -- 1.2 \\ 
 \hline
 Horizontal shift & 0 -- 0.15 \\ 
 \hline
 Vertical shift & 0 -- 0.05 \\ 
 \hline
 Brightness & -20 -- 20 \\ 
 \hline
 Contrast & 0.7 -- 2 \\ 
 \hline
 Saturation & 0.75 -- 1.25 \\ 
 \hline
 Hue & -0.1 -- 0.1 \\ 
 \hline
\end{tabular}

\vspace*{.5 cm}
Source: Author.
\end{table}

\begin{figure}[!ht]
    \centering
    \caption{Sample frames of face and hands in a video with random data augmentation.}
    \includegraphics[width=.99\textwidth]{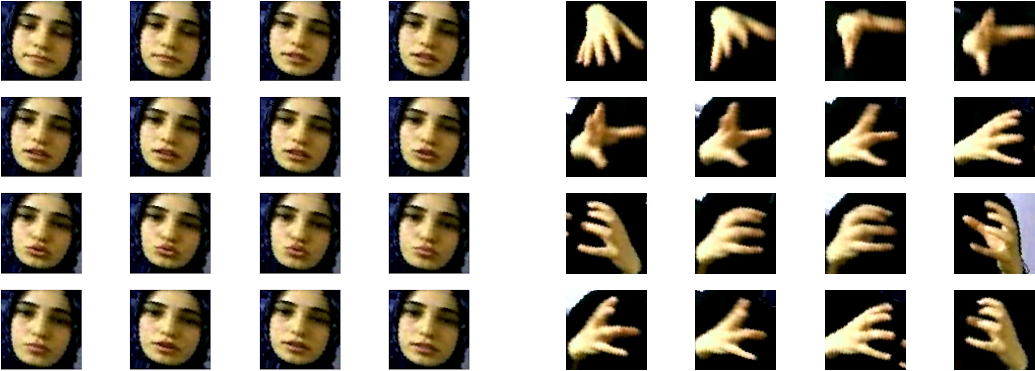}
    \label{fig:augmentation_crop_sample}

    Source: Author.
\end{figure}

The intensity operations, on the other hand, involved changes in brightness, contrast, hue and saturation of the image. Figure \ref{fig:augmentation_crop_sample} displays an example of applying random data augmentation to one of the videos in the AUTSL database. As observed, the same operations are applied to all video frames.

\subsection{Handcrafted features} \label{handcrafted features}
As shown in Figure \ref{fig:language_params}, the majority of the image is used to represent the location of the hands. However, this linguistic parameter incurs computational costs, once it is necessary to process all pixels within the region. Additionally, it contributes to the overall complexity of the system, as it requires accounting for invariance to scenery and lighting conditions.

\begin{figure}[!ht]
    \centering
    \caption{Triangular figure generated by connecting the centroids of the bounding boxes of the face (fc), right hand (rh), and left hand (lh).}
    \includegraphics[width=.35\textwidth]{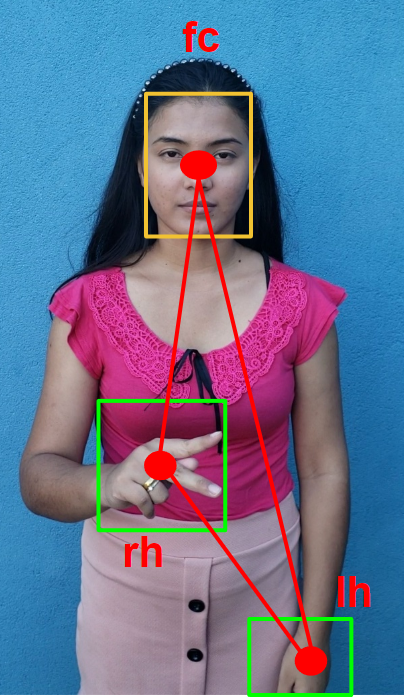}
    \label{fig:triangular_figure}

    Source: Author.
\end{figure}

Based on that, our research proposes handcrafted features to represent location and movement information in a simplified manner. This is achieved by leveraging spatial information derived from the centroid positions of the bounding boxes inferred by the object detector. To accomplish this, the initial step involves determining the coordinates $(x, y)$ of the central region within the figures that delineate the location of the hands and the face, as represented in Equation \ref{eq:centroid}:

\begin{equation} \label{eq:centroid}
   c(x_{min}, x_{max}, y_{min}, y_{max}) = [\frac{x_{min} + x_{max}}{{2}}\space,\space \frac{y_{min} + y_{max}}{{2}}],
\end{equation}

\noindent where $x_{min}$ and $x_{max}$ represent the minimum and maximum points of the bounding box on the x-axis, respectively, and $ y_{min}$ and $y_{max}$ are the minimum and maximum points on the y-axis. Once the centroids are determined, they can be connected to form a triangular shape, as illustrated in Figure \ref{fig:triangular_figure}. This geometric figure provides a spatial reference for the position of the hands in relation to the face of the interpreter, which remains static during sign execution.

That said, we created a vector that numerically represents the triangle through its properties by calculating the following features:

\begin{itemize}
    \item Triangle sides;
    \item External and internal angles;
    \item Perimeter and semi-perimeter;
    \item Area;
    \item Height;
    \item Hand movements.
\end{itemize}

The first calculated feature was the distance of the centroids in the space, based on the Equation \ref{eq:euclidian}:

\begin{equation} \label{eq:euclidian}
   dist(a, b) = \sqrt{(b_{x} - a_{x})^2 + (b_{y} - a_{y})^2},
\end{equation}

\noindent considering that $x$ and $y$ represent the 2D image axes and $a$ and $b$ identify the centroid coordinates. Therefore, the vector with the hand and face distances can be represented as follows: $\mathbf{d} =[dist(rh, lh), dist(rh, fc), dist(lh, fc)]$, where $rh, lh,$ and $fc$ are the centroids coordinates of the right hand, left hand, and face of the interpreter, respectively. By obtaining the side lengths of the triangle through the relative distance of the three centroids, it is possible to calculate the perimeter ($pr$) and semi-perimeter ($spr$), given by Equations \ref{eq:perimetro} and \ref{eq:semiperimetro}, respectively:

\begin{equation} \label{eq:perimetro}
   pr(\mathbf{d}) =  \sum \left({d_i}\right),
\end{equation}

\begin{equation} \label{eq:semiperimetro}
   spr(pr) = \frac{pr}{{2}}.
\end{equation}

In contrast, the calculation of the area varies depending on the type of triangle (equilateral, isosceles, scalene, etc.). However, since several distinct figures can be formed during sign execution, Heron's formula \cite{benyi200387} was adopted to generically calculate the area ($\lambda$), represented in Equation \ref{eq:area}:

\begin{equation} \label{eq:area}
   \lambda = \sqrt{spr (spr - d_{1}) (spr - d_{2}) (spr - d_{3}) }.
\end{equation}

The height, on the other hand, is given by Equation \ref{eq:altura}:

\begin{equation} \label{eq:altura}
   h(\lambda, \mathbf{d}) = \frac{{2} \lambda}{d_{1} + \gamma},
\end{equation}

\noindent in which the distance between the right hand and the left hand ($d_{1}$) is utilized as the base of the triangle and $\gamma$ represents a small positive constant value (1e-10) used to avoid division by zero. In addition, the calculation of the internal angles of the triangle is given by applying the cosine law \cite{robusto1957cosine}, represented in Equation \ref{eq:cosine_theorem}:

\begin{equation} \label{eq:cosine_theorem}
   op^2 = adj_1^2 + adj_2^2 - 2adj_1adj_2\cos \alpha,
\end{equation}

\noindent in this case, $op$ represents the size of the side opposite to the angle $\alpha$, while $adj_1$ and $adj_2$ are the adjacent sides. Once this is done, the three external angles can be easily calculated by the sum of the internal angles not adjacent to each external angle, as stated in Equation \ref{eq:external_angle}:

\begin{equation} \label{eq:external_angle}
   \hat{\alpha} = \alpha_a + \alpha_b,
\end{equation}

\noindent where $\alpha_a$ e $\alpha_b$ are internal angles not adjacent to the external angle $\hat{\alpha}$. Finally, the hand movement is calculated based on the difference between the positions of each hand at different time intervals, as explained in Section \ref{video preprocessing}, providing insights into the direction, magnitude, and pattern of hand movements during sign execution.

\begin{figure}[!ht]
    \centering
    \caption{Comparison between the shape and values of the triangle for the same sign captured in different ways. The internal angles are represented in blue, while the distances are shown in orange.}
    \includegraphics[width=.99\textwidth]{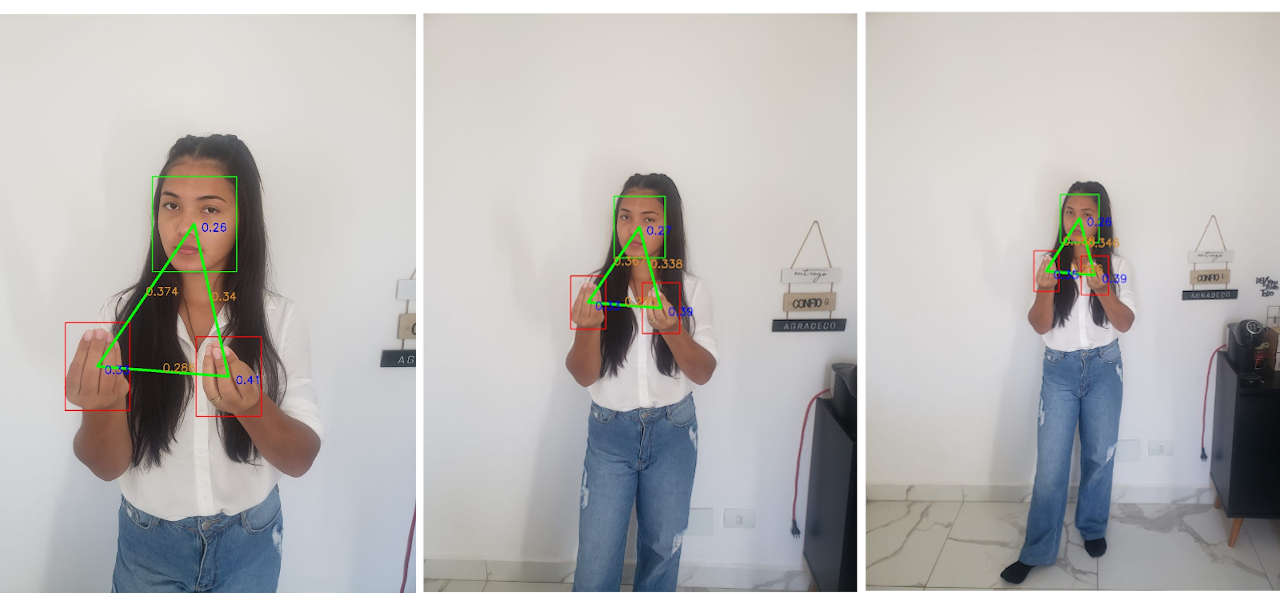}
    \label{fig:triangle_comp}

    Source: Author.
\end{figure}

One of the challenges in representing the triangle is to ensure that similar figures have corresponding values, regardless of the interpreter's distance from the camera. To achieve this, we normalized the features by dividing the distances of each vertex by the perimeter of the triangle, as represented in Equation \ref{eq:distance_norm}:

\begin{equation} \label{eq:distance_norm}
   \hat{\mathbf{d}}(\mathbf{d}, spr) =  \left[\frac{distance}{pr} \mid distance \in \mathbf{d} \right],
\end{equation}

\noindent ensuring that all the features maintain approximately the same proportion. This guarantees that the feature vector accurately represents the shape of the figure, without being influenced by variations in video recording, as illustrated in Figure \ref{fig:triangle_comp}.

The values of the internal angles (in blue) are: [0.33, 0.41, 0.26], [0.33, 0.39, 0.27], [0.35, 0.39, 0.26], from the closest to the farthest image, respectively, while the distances between the vertices, following the same order, are: [0.286, 0.34, 0.374], [0.295, 0.338, 0.367], [0.286, 0.346, 0.368]. This clearly indicates that the features of the figure are primarily related to the shape it assumes, rather than external factors of the image, even considering the variations caused by the stochasticity of the object detector's inference, which may slightly displace the centroid of the bounding boxes.

Finally, the perimeter and semi-perimeter values were excluded from the feature vector because they assumed fixed values after normalization. With this adjustment, a total of 13 distinct features were derived from the triangle, as highlighted in Table \ref{table:triangle_features}, resulting in a vector that was used as one of the classifier inputs. By adopting this approach, we mitigate the margins for ambiguities arising from the lack of discriminative features. Additionally, we reduce the computational costs, as less relevant areas of the image, such as the background, are replaced by the numerical representations that are simpler to process.

\begin{table}[h!]
\centering
\caption{Features extracted from the triangle figure.}
\label{table:triangle_features}
\vspace*{.25 cm}

 \begin{tabular}{| c c|} 
 \hline
 \textbf{Feature name} & \textbf{Number of features} \\ [0.5ex] 
 \hline\hline
 Triangle sides & 3 \\ 
 \hline
 Internal angles & 3 \\ 
 \hline
 External angles & 3 \\ 
 \hline
 Height & 1 \\ 
 \hline
 Area & 1 \\ 
 \hline
 Hand movements & 2 \\ 
 \hline
\end{tabular}

\vspace*{.5 cm}
Source: Author.
\end{table}

\subsection{Triangle figure} \label{triangle figure}
As mentioned in Section \ref{related_works}, the work by Xiao et al. \cite{xiao2020multi} is one of the most closely related to our research. Therefore, we reproduced the technique proposed by the authors to facilitate a comparison between the usage of the triangle figure and the handcrafted feature vector proposed in Section \ref{handcrafted features}. Nonetheless, certain modifications were implemented to add more semantic value to the figure. Firstly, we treated the background as an absence of information by padding the image with zeros, reducing the amount of information to be processed. Additionally, we filled the area of the triangle with the color white, creating a distinction from the background, while the colors green, red, and blue were used to identify the face, and hands.

\begin{figure}[ht]
    \centering
    \caption{Triangle figure generated from the sign execution.}
    \includegraphics[width=.99\textwidth]{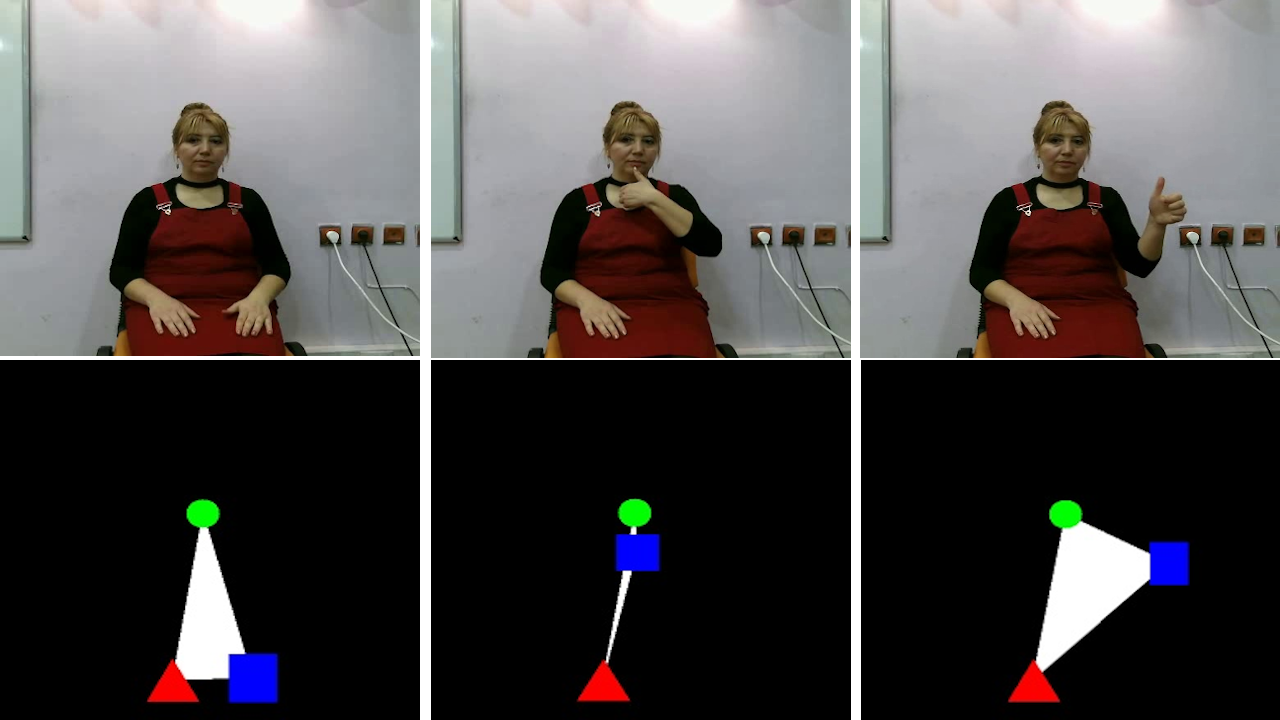}
    \label{fig:triangle_fig}

    Source: Author.
\end{figure}

Moreover, the geometric shapes of the original work were maintained, using a circle for the face and a triangle or a square for each hand. However, we differentiated the shapes by assigning the square to the dominant hand and the triangle to the non-dominant hand, following the criteria defined in Section \ref{video preprocessing}. Figure \ref{fig:triangle_fig} illustrates the resulting triangle image.

As observed, the generated figure distinguishes the elements present in the image not only by their shape but also by the color of the objects. In contrast, Xiao et al. \cite{xiao2020multi} used the same colors for all figures, and only delineated the area of the triangle with small red borders. Finally, the triangular figures were resized to a resolution of 128x128 pixels and employed as one of the information channels for sign classification, as detailed in Section \ref{classification system triangle fig}.

\subsection{Classification system using handcrafted features} \label{classification system handcrafted}
The first step to create the architecture illustrated in Figure \ref{fig:full_architecture} was to resize all the input segments to a resolution of 100x100 pixels. However, since the left and right hands present similar information, they were horizontally joined, keeping the dominant hand on the same side to ensure consistency across the frames of the video. This resulted in a single image with a resolution of 100x200 pixels. The EfficientNetB0 architecture \cite{tan2019efficientnet} was then employed as the feature extractor due to its excellent trade-off between the number of parameters and accuracy.

Regarding the extraction of the temporal patterns of each input channel, different RNNs models were adopted. Even so, the selection of recurrent cell type, normalization, number of neurons, and hidden layers is heavily dependent on the problem domain \cite{geron2019hands}, typically involving empirical trial and error to find the optimal neural network architecture for the task. Furthermore, several methods exist to merge model channels to better exploit the available information for classification. Based on that, we adopted the Keras Tuner library \cite{omalley2019kerastuner} in order to facilitate the search for the best combination of hyperparameters.

For the proper functioning of the tool, it is necessary to define a search space with reasonable minimum and maximum values for each hyperparameter, considering the problem context. This includes specifying possible types of neurons, fusion methods, recurrent cells, among other architectural parameters to be tested. In our work, to create the three different RNNs we used LSTM \cite{hochreiter1997long} or GRU \cite{chung2014empirical} type cells, which can be bidirectional \cite{schuster1997bidirectional} and contain an attention layer \cite{bahdanau2014neural}. Furthermore, the ranges for the number of layers, neurons and Dropout \cite{srivastava2014dropout} probability varied for each neural network, as shown in Table \ref{table:search_space}.

\begin{table}[h!]
\centering
\caption{Variation in hyperparameter ranges for each RNN channel.}
\label{table:search_space}
\vspace*{.25 cm}

 \begin{tabular}{| c c c c|} 
 \hline
 \textbf{} & \textbf{Face} & \textbf{Hands} & \textbf{Triangle features} \\ [0.5ex] 
 \hline\hline
 Minimum of layers & 1  & 1 & 1 \\ 
 \hline
 Maximum of layers & 3  & 5 & 3 \\ 
 \hline
 Minimum of neurons & 64 & 32 & 64\\ 
 \hline
 Maximum of neurons & 448 & 512 & 320\\ 
 \hline
 Minimum of \textit{Dropout} & 5\% & 5\%  & 5\% \\ 
 \hline
 Maximum of \textit{Dropout} & 40\% & 40\% & 40\%\\ 
 \hline
\end{tabular}

\vspace*{.5 cm}
Source: Author.
\end{table}

Furthermore, three approaches for merging the features extracted by the RNN channels were tested to generate the final classification. First, the early fusion was adopted, concatenating the three feature vectors into a single vector of higher dimensionality, which was then submitted to a Softmax classifier. In the late fusion, on the other hand, each vector was individually classified by the Softmax layer, and the average of the probabilities was used to achieve the final predictions. Lastly, a meta-learning approach was employed, combining the feature vectors and submitting them to a new multi-layer Perceptron network. This network comprises one to two layers with 64 to 512 neurons and a Dropout rate varying between 5\% to 40\%.

To adjust the hyperparameters with the aforementioned settings, the Hyperband algorithm \cite{lisha2018bandit} was used in Keras Tuner, aiming to balance the exploration of the search space and the execution time. Moreover, each RNN was individually optimized to ensure that the results were the best possible for processing the information from the specific channel, without being influenced by other channels.

\subsection{Classification system using triangle figure} \label{classification system triangle fig}
The features of the triangle figure, on the other hand, were extracted using the MobileNetV2 architecture \cite{sandler2018mobilenetv2}, which has a small number of parameters. This decision was driven by the simplicity of the triangle image, which displays easily discernible patterns comprised of only a few colors and geometric shapes. Besides that, it is noteworthy that the initial training utilized the EfficientNetB0 model, however, since the accuracy achieved was comparable to that of MobileNetV2, the selection of the simpler model was justified.

After obtaining the feature vector, a RNN was also optimized for this channel using Keras Tuner, according to the hyperparameters presented in Table \ref{table:search_space_triangle_fig}. Finally, a series of models were trained considering the combination with the information from the triangle figure, allowing for a direct comparison to the method proposed by this research.

\begin{table}[h!]
\centering
\caption{Hyperparameter ranges for the RNN classifier used to the triangle figure.}
\label{table:search_space_triangle_fig}
\vspace*{.25 cm}

 \begin{tabular}{| c c|} 
 \hline
 \textbf{} & \textbf{Triangle figure} \\ [0.5ex] 
 \hline\hline
 Minimum of layers & 1 \\ 
 \hline
 Maximum of layers & 4 \\ 
 \hline
 Minimum of neurons & 32 \\ 
 \hline
 Maximum of neurons & 512\\ 
 \hline
 Minimum of \textit{Dropout} & 5\% \\ 
 \hline
 Maximum of \textit{Dropout} & 40\%\\ 
 \hline
\end{tabular}

\vspace*{.5 cm}
Source: Author.
\end{table}

\subsection{Experimental methodology} \label{experimental methodology}
The main experiment conducted in this research aims to investigate the impact of the triangle feature vector on classification accuracy, demonstrating its ability to represent the location and movement of the sign. For this purpose, after the hyperparameter optimization, 7 distinct classifiers were trained using different combinations of the input channels.

Each training was conducted for a total of 40 epochs, with a batch size of 30, an initial learning rate of 0.001, and using the Adam optimization function \cite{kingma2014adam}. However, a multi-phase methodology \cite{sarhan2022multi} was implemented for the channels of the hands and face, ensuring a proper fine-tuning of the CNNs.

The first phase used the aforementioned parameters to warm up the RNN model, freezing all the CNN weights transferred from the training of the ImageNet \cite{deng2009imagenet} dataset. Four additional training phases were conducted to adapt the CNNs to the problem domain, unfreezing layers 221, 162, 119, and 75 of EfficientNetB0, with 40, 35, 35, and 30 epochs, respectively. In these phases, we utilized the SGD optimizer \cite{geron2019hands}, as it produced better results than Adam for fine-tuning, and the learning rate ranged from 0.00001 to 0.001 through a triangular decay \cite{goyal2017accurate}. All the fine-tuned CNNs were used in the training of the combined channels.

Furthermore, another four classifiers were trained by considering the combination of the triangle figure with the other information channels, serving as a comparison between the technique proposed by this research and those present in the literature. The training followed the same parameters mentioned above for the other channels, but the fine-tuning of the CNN was performed through a single step after warming up the model, unfreezing the layer 143 of MobileNetV2, as unfreezing more layers led to degradation of the results.

Finally, to evaluate the results, accuracy values on the test set were obtained for each of the models, along with the inference time on CPU (Core i5-10400), GPU (RTX 3060), and the number of parameters. The inference time in milliseconds was obtained by averaging the model's inference time for each of the videos (with 16 frames) in the test set of the AUTSL dataset. Together with the number of parameters, these metrics represent the complexity and computational cost of the classifier. The models and the source code developed in this research are publicly available on GitHub \footnote{\url{https://github.com/AlvaroCavalcante/multi-cue-sign-language}}.

\section{Results}
\label{results}

Regarding the results, the first step involved optimizing the hyperparameters of each information channel to find the best architectures for feature extraction, generating RNNs with the specifications shown in Table \ref{table:results_keras_tuner}. As observed, most models achieved better convergence with fewer than three layers and with the adoption of GRU cells and attention layers, highlighting the effectiveness of these techniques for this problem. Furthermore, the early fusion approach was selected to combine the channels, as it proved to be more effective for the task. Once the architectures were defined, it became possible to train the different models for sign classification, leading to the results presented in Table \ref{table:resultados_autsl}.

\begingroup
\setlength{\tabcolsep}{5pt} 
\renewcommand{\arraystretch}{1} 
\begin{table}[!ht]
\centering
\caption{Best RNN architectures found for each information channel using Keras Tuner.}
\label{table:results_keras_tuner}
\vspace*{.25 cm}

 \begin{tabular}{|c c c c c|} 
 \hline
 \textbf{Hyperparameters} & \textbf{Hands} & \textbf{Face} & \textbf{Triangle} & \textbf{Triangle Figure} \\ [0.5ex] 
 \hline\hline
    Bidirectional & Yes & No & No & Yes \\ 
  \hline
    Attention & Yes & Yes & Yes & Yes \\
  \hline
    Recurent cell & GRU & GRU & GRU & GRU \\ 
\hline
    Dropout & 25\% & 5\% & 10\% & 25\% \\ 
 \hline
    Neurons of layer 1 & 288 & 384 & 320 & 288 \\ 
  \hline
    Neurons of layer 2 & - & 320 & 192 & 416 \\ 
  \hline
    Neurons of layer 3 & - & 192 & - & - \\ 
 \hline

\hline
\end{tabular}

\vspace*{.5 cm}
Source: Author.
\end{table}
\endgroup

\begingroup
\setlength{\tabcolsep}{5pt} 
\renewcommand{\arraystretch}{1.5} 
\begin{table}[!ht]
\centering
\caption{Comparison of model results on the AUTSL dataset.}
\label{table:resultados_autsl}
\vspace*{.25 cm}
\resizebox{\textwidth}{!}{

 \begin{tabular}{|c c c c c|} 
 \hline
 \textbf{Input channel} & \textbf{\makecell{Test \\accuracy}} & \textbf{\makecell{Inference \\ on GPU (ms)}} & \textbf{\makecell{Inference \\ on CPU (ms)}} & \textbf{Parameters} \\ [0.5ex] 
 \hline\hline
    Hands & 71.83\% & 42 & 110 & 6,893,525 \\ 
  \hline
    Triangle & 46.85\% & \textbf{28} & \textbf{28} & \textbf{661,490} \\
  \hline
    Triangle Figure & 54.96\% & 37 & 69 & 7,641,074 \\
  \hline
    Face & 24.49\% & 38 & 69 & 6,986,453 \\ 
  \hline
    Hands + Face & 68.88\% & 54 & 153 & 13,879,752 \\ 
  \hline
    Hands + Triangle & 79.79\% & 44 & 118 & 7,554,789 \\ 
  \hline
    Hands + Triangle Figure & 76.06\% & 53 & 154 & 14,534,373 \\ 
  \hline
    Face + Triangle & 54.69\% & 41 & 70 & 7,647,717 \\ 
  \hline
    Face + Triangle Figure & 55.74\% & 52 & 114 & 14,627,301 \\ 
 \hline
    Hands + Face + Triangle & \textbf{80.72\%} & 55 & 153 & 14,541,016 \\ 
  \hline
    Hands + Face + Triangle Figure & 74.72\% & 65 & 199 & 21,520,600 \\ 
 \hline

\hline
\end{tabular}}

\vspace*{.5 cm}
Source: Author.
\end{table}
\endgroup

With respect to the performance of the individual channels, the hands resulted in better accuracy for discriminating the signs, as it contains the highest number of linguistic parameters. In contrast, the face alone was unable to effectively distinguish the signs as it relies solely on facial expressions. The triangle channel, even though composed of only 13 numerical values, can correctly classify nearly half of the test samples, highlighting the success of the handcrafted features in representing the movement and location of the hands. Moreover, the triangle figure performed better individually than the simple feature vector, increasing the test accuracy by 8.11\%.

The fusion of channels allows for a more precise assessment of how each feature contributes to the discrimination of signs, as channels with similar features show marginal accuracy gains when concatenated. Based on that, it becomes evident that the features from the hands and the triangle features complement each other in representing the linguistic parameters, leading to an increase in accuracy of approximately 8\%. 

\begin{figure}[!ht]
    \centering
    \caption{Visual intersection between the channels of hands and face.}
    \includegraphics[width=.90\textwidth]{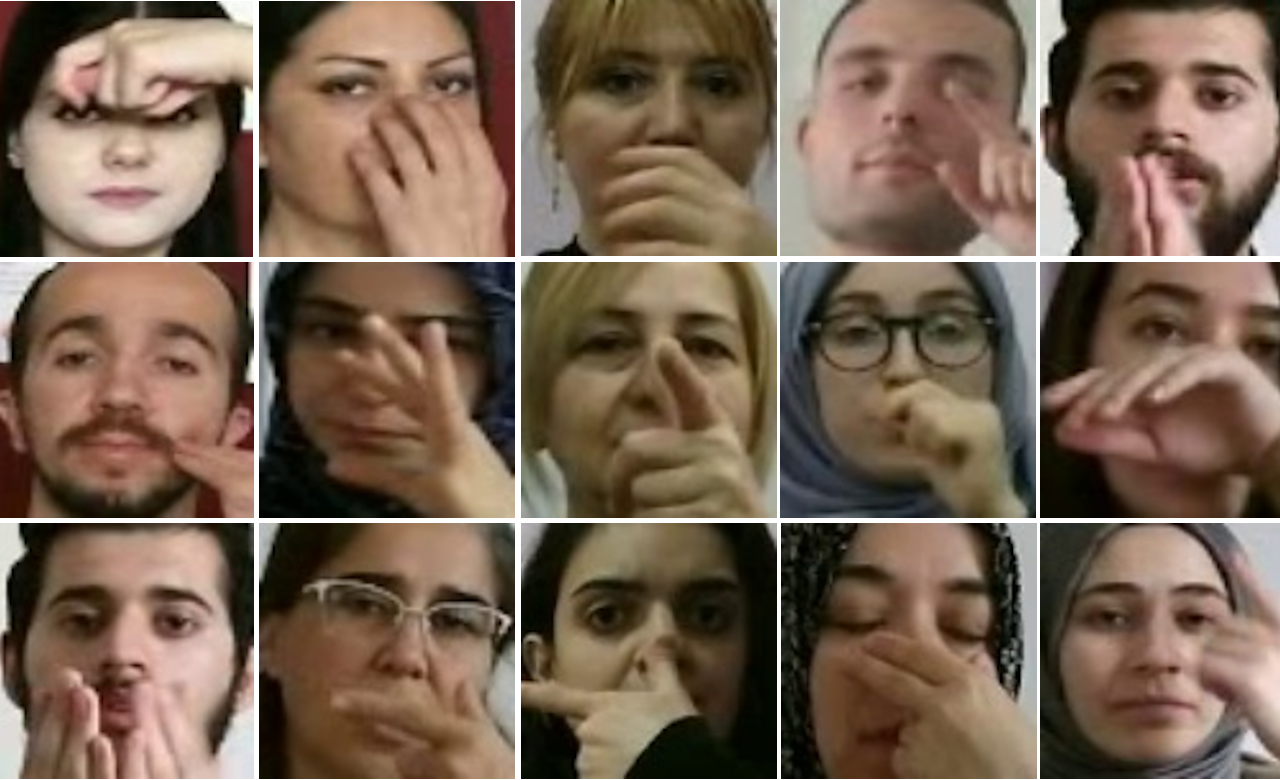}
    \label{fig:face_hand_intersec}

    Source: Author.
\end{figure}

Although fusing the triangle data with the face also led to improved results, this enhancement can likely be attributed to the overlap between the hand and face channels. This occurs because, in some signs, the hands are near or overlapping the interpreter's face, thus influencing the features extracted by the CNN, as shown in Figure \ref{fig:face_hand_intersec}. Combining all three data channels (hands, face, and triangle features) resulted in a marginal accuracy increase of less than 1\%, indicating the lack of impact from the facial information, not only due to the aforementioned overlap but also due to the limited number of signs with accentuated facial expressions in this dataset.

Contrary to what was observed individually, the performance of the triangle figure was not superior to the triangle features when combined with the other channels, indicating the lower effectiveness of the geometric figure image in precisely representing linguistic parameters, as well as the challenge of concatenating features extracted by more complex deep learning models. Additionally, there is a noticeable increase in the number of parameters and inference time for models when incorporating this channel, even with the adoption of MobileNetV2 (2,257,984 million parameters) compared to EfficientNetB0 (4,049,571 million parameters) for extracting features from the geometric figure.

That said, although the triangle figure resulted in a 4.23\% increase in accuracy when combined with the hands channel, the additional computational cost may not justify adopting this approach, especially considering the superiority of using handcrafted features, which achieved an accuracy 3.73\% higher. Moreover, the difference in the number of parameters and inference time of each channel highlights that efficient exploration of linguistic parameters allows for achieving good results without the need for overly complex models, which is of utmost importance in the area of sign language recognition.

Based on these findings, the results on the AUTSL dataset validate the technique developed in this research for representing the location and movement of the hands through a simple feature vector, achieving over 80\% accuracy on a challenging dataset with diverse interpreters, backgrounds, and lighting conditions. While other state-of-the-art works have achieved better accuracy \cite{de2021isolated, gruber2021mutual, sincan2021using}, most of them utilize complex architectures, involving body pose extraction methods and combining predictions from various distinct models. In contrast, the technique developed in this research achieved a 7.96\% increase in accuracy with fewer than 670 thousand parameters and incurred only 2 and 8 milliseconds of additional inference time on the GPU and CPU, respectively.

\section{Conclusion and future work}
\label{conclusion}

In this research, a handcrafted feature vector was proposed to represent the linguistic parameters that comprise the signs. Using the spatial references provided by the bounding box centroids, we created a triangular shape by interconnecting the interpreter's hands and face. Subsequently, a feature vector was obtained to numerically represent the geometric figure. The adoption of these features, along with several pre-processing techniques, resulted in an efficient sign language recognition system with low computational cost. Specifically, the proposed features increase the accuracy by 7.96\%, while requiring fewer than 670 thousand parameters and less than 10 milliseconds of additional inference time.

Despite promising results, we noticed a limitation in our technique for representing image depth. Consequently, the model encountered difficulties in distinguishing signs where the interpreter's hands remain in the same spatial coordinates but move back and forth. Besides that, the final model was not tested on portable devices, indicating that the solution is not yet accessible for deaf people to use in their daily routines.

In future work, we aim to enhance and extract novel features from the spatial references to provide a more accurate representation of linguistic parameters, with a specific emphasis on capturing depth movement. Moreover, it is important to evaluate the system's inference time on smartphones, thereby presenting a scenario that closely resembles real-world usage.

Finally, the triangle features will be used to compose different model architectures, including for the problem of continuous sign language recognition, further showing the effectiveness of the technique in different datasets and contexts.

\section*{Funding sources}
\label{funding}
This research did not receive any specific grant from funding agencies in the public, commercial, or not-for-profit sectors.



\bibliographystyle{elsarticle-num} 
\bibliography{ref}


\end{document}